%% file: main.tex
\documentclass[letterpaper, 10 pt, journal, twoside]{ieeetran}  % Comment this line out if you need a4paper

\IEEEoverridecommandlockouts                              % This command is only needed if 
\usepackage{graphicx}
\usepackage{amsmath} % assumes amsmath package installed
\usepackage{amsthm}
\newtheorem{assumption}{Assumption}
\newtheorem{theorem}{Theorem}
\newtheorem{remark}{Remark}
\usepackage{hyperref}
\usepackage{amssymb}  % assumes amsmath package installed
\usepackage{algorithm}
\usepackage{algpseudocode}
\algnewcommand\algorithmicinput{\textbf{INPUT:}}
\algnewcommand\INPUT{\item[\algorithmicinput]}
\algnewcommand\algorithmicoutput{\textbf{OUTPUT:}}
\algnewcommand\OUTPUT{\item[\algorithmicoutput]}
\usepackage{subfiles}

\usepackage{enumitem}
\usepackage{color}

\usepackage[font=footnotesize,belowskip=-10pt,aboveskip=3pt]{caption}
\renewcommand{\baselinestretch}{0.92}

\title{
RAT iLQR: A Risk Auto-Tuning Controller to Optimally Account for Stochastic Model Mismatch
}
\author{Haruki Nishimura$^{1}$, Negar Mehr$^{2}$, Adrien Gaidon$^{3}$, and Mac Schwager$^{1}$% <-this % stops a space
\thanks{Manuscript received: 10, 15, 2020; Revised 12, 25, 2020; Accepted 12, 29, 2020.}%Use only for final RAL version
\thanks{This paper was recommended for publication by Editor Lucia Pallottino upon evaluation of the Associate Editor and Reviewers' comments. 
Toyota Research Institute (``TRI")  provided funds to assist the authors with their research but this article solely reflects the opinions and conclusions of its authors and not TRI or any other Toyota entity. This work was also supported in part by ONR grant number N00014-18-1-2830, a JASSO fellowship, and a Masason Foundation fellowship. The authors are grateful for this support.}% <-this % stops a space
\thanks{$^{1}$Haruki Nishimura and Mac Schwager are with the Department of Aeronautics and Astronautics, Stanford University, Stanford, CA 94305 USA
        {\tt\small  \{hnishimura, schwager\}@stanford.edu}}%
\thanks{$^{2}$Negar Mehr is with the Aerospace Engineering Department, the University of Illinois at Urbana-Champaign, Urbana, IL 61801 USA
        {\tt\small negar@illinois.edu}}
\thanks{$^{3}$Adrien Gaidon is with Toyota Research Institute, Los Altos, CA 94022 USA
        {\tt\small adrien.gaidon@tri.global}}%
\thanks{Digital Object Identifier (DOI): 10.1109/LRA.2020.3048660}
}

\begin{document}
\maketitle
%\thispagestyle{empty}
%\pagestyle{empty}
%%%%%%%%%%%%%%%%%%%%%%%%%%%%%%%%%%%%%%%%%%%%%%%%%%%%%%%%%%%%%%%%%%%%%%%%%%%%%%%%
\begin{abstract}
% Do we want to start with what is the hard thing that we are actually solving here?
%This paper proposes DRAKE, a novel MPC algorithm for distributionally robust control of nonlinear stochastic systems under a given KL divergence bound between the model and the true distributions. Our approach relies on the fundamental equivalence between distributionally robust control and risk-sensitive optimal control. A locally optimal solution to the resulting bilevel optimization is derived with the aid of recent advancements in nonlinear risk-sensitive optimal control. We show the effectiveness of our framework in a collision avoidance scenario with erroneous human motion prediction. We also demonstrate DRAKE's capability to dynamically adjust the risk-sensitivity parameter online, which overcomes a limitation of existing risk-sensitive optimal control methods.
% How about writing the abstract with more info, For instance start with something like reasoning about model errors being key for robots' success, then talk about our control algorithm enabling the control algorithm reason about the model mismatches. Then talk about the connection between risk-sensitive control. Talk about the experiments. Then, add more elaboration on the fact that due to the connection to risk-sensitive control, our framework also provide an algorithm for auto-adjusting risk-sensitivity of risk-sensitive control which has been one of the main difficulties of risk-sensitive control. 
Successful robotic operation in stochastic environments relies on accurate characterization of the underlying probability distributions, yet this is often imperfect due to limited knowledge. This work presents a control algorithm that is capable of handling such distributional mismatches. Specifically, we propose a novel nonlinear MPC for distributionally robust control, which plans locally optimal feedback policies against a worst-case distribution within a given KL divergence bound from a Gaussian distribution. Leveraging mathematical equivalence between distributionally robust control and risk-sensitive optimal control, our framework also provides an algorithm to dynamically adjust the risk-sensitivity level online for risk-sensitive control. The benefits of the distributional robustness as well as the automatic risk-sensitivity adjustment are demonstrated in a dynamic collision avoidance scenario where the predictive distribution of human motion is erroneous.
 \end{abstract}
%%%%%%%%%%%%%%%%%%%%%%%%%%%%%%%%%%%%%%%%%%%%%%%%%%%%%%%%%%%%%%%%%%%%%%%%%%%%%%%%


\section{INTRODUCTION}
\label{sec: intro}
\subfile{introduction}

\section{RELATED WORK}
\label{sec: related}
\subfile{related_work}

\section{PROBLEM STATEMENT}
\label{sec: problem}
\subfile{problem_statement}

\section{RAT iLQR ALGORITHM}
\label{sec: algorithm}
\subfile{algorithm}

\section{RESULTS}
\label{sec: results}
\subfile{results}

\section{CONCLUSIONS}
\label{sec: conclusions}
\subfile{conclusions}

\bibliographystyle{IEEEtran}
\bibliography{IEEEabrv, references}

\end{document}

%% file: introduction.tex
\IEEEPARstart{P}{roper} modeling of a stochastic system of interest is a key step towards successful control and decision making under uncertainty. 
In particular, accurate characterization of the underlying probability distribution is crucial, as it encodes how we expect the system to behave \textit{unexpectedly} over time. 
However, such a modeling process can pose significant challenges in real-world problems. On the one hand, we may have only limited knowledge of the underlying system, which would force us to use an %a simpler, 
erroneous model. On the other hand, even if we can perfectly model a complicated stochastic phenomenon, such as a complex multi-modal distribution, %it may still not be usable by the control or planning algorithm of our choice.
it may still not be appropriate for the sake of real-time control or planning.
Indeed, many model-based stochastic control methods require a Gaussian noise assumption, and many of the others need computationally intensive sampling. 

%The present work addresses this problem via distributionally robust, risk-sensitive control. Specifically, we propose DRAKE, a novel model predictive control (MPC) algorithm for nonlinear, non-Gaussian systems with non-convex costs. Our control algorithm accounts for a potential distributional mismatch between a Gaussian process noise and the true model within a certain Kullback-Leibler (KL) divergence bound. Our contribution is built on the mathematical equivalence between distributionally robust control and risk-sensitive optimal control~\cite{petersen2000minimax}. Risk-sensitive optimal control seeks to optimize the entropic risk measure \cite{majumdar2020should}:
The present work addresses this problem via distributionally robust control, wherein a potential distributional mismatch is considered between a baseline Gaussian process noise and the true, unknown model within a certain Kullback-Leibler (KL) divergence bound. The use of the Gaussian distribution is advantageous to retain computational tractability without the need for sampling in the state space. Our contribution is a novel model predictive control (MPC) method for nonlinear, non-Gaussian systems with non-convex costs. This controller would be useful, for example, to safely navigate a robot among human pedestrians while the stochastic transition model for humans is not perfect.

It is important to note that our contribution is built on the mathematical equivalence between distributionally robust control and risk-sensitive optimal control~\cite{petersen2000minimax}. Unlike the conventional stochastic optimal control that is concerned with the expected cost, risk-sensitive optimal control seeks to optimize the following entropic risk measure \cite{majumdar2020should}:
\begin{align}
    R_{p,\theta}(J) \triangleq \frac{1}{\theta} \log\mathbb{E}_p\left[\exp(\theta J)\right],
\end{align}
%where $p$ is a probability distribution, 
where $p$ is a probability distribution characterizing any source of randomness in the system,
$\theta > 0$ is a user-defined scalar parameter called the risk-sensitivity parameter, and $J$ is an optimal control cost. The risk-sensitivity parameter $\theta$ determines a relative weight between the expected cost and other higher-order moments such as the variance \cite{whittle2002risk}. Loosely speaking, the larger $\theta$ becomes, the more the objective cares about the variance and is thus more risk-sensitive. %The entropic risk has been extensively studied in control theory \cite{whittle1981risk, glover1988relations}, especially for linear Gaussian systems where a closed-form dynamic programming (DP) solution exists. 

\begin{figure}[t]
    \begin{center}
	\begin{tabular}{c}
		\begin{minipage}[t]{0.46\columnwidth}
			\centering
 			\scalebox{1.0}[1.0]{\includegraphics[trim=5 0 5 0,clip,width=1.0\columnwidth]{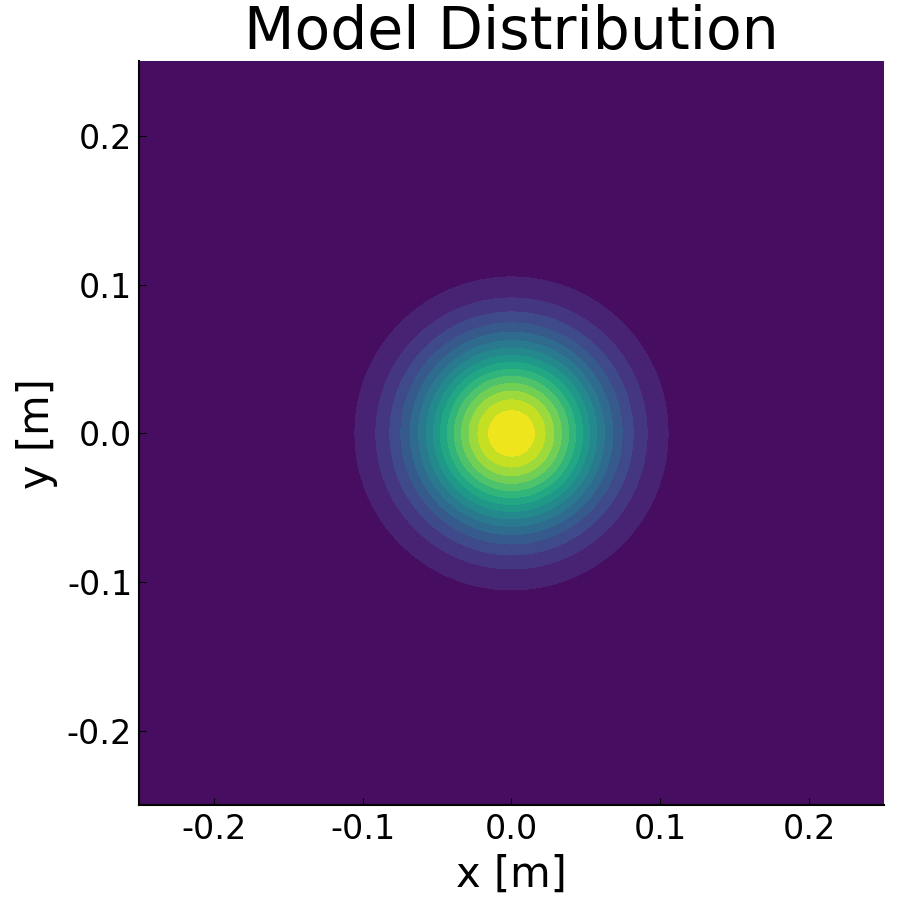}}
		\end{minipage}
		\begin{minipage}[t]{0.46\columnwidth}
			\centering
			\scalebox{1.0}[1.0]{\includegraphics[trim=5 0 5 0,clip,width=1.0\columnwidth]{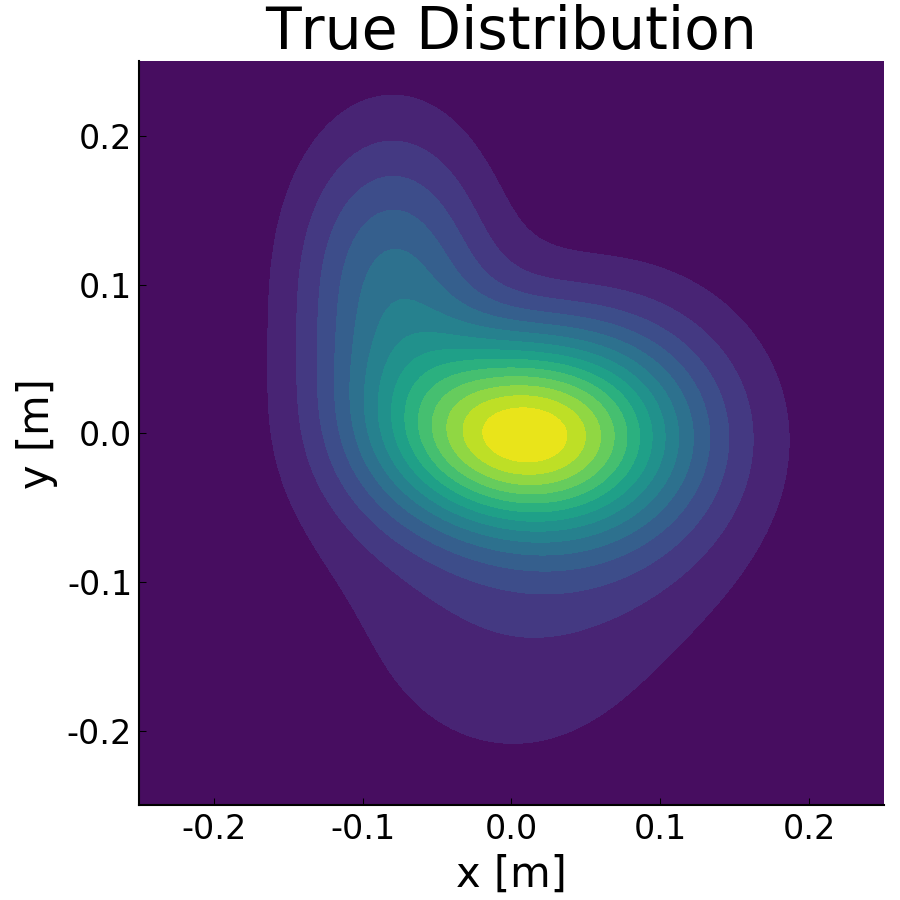}}
		\end{minipage}
	\end{tabular}
    \caption{Model-based stochastic control methods often require a Gaussian noise assumption, such as the one in the left that represents process noise in pedestrian motion under a collision avoidance scenario (see Section \ref{sec: results}). However, the true stochastic model can be highly multi-modal and better captured by a more complex distribution as shown in the right, which we may not exactly know. The proposed MPC effectively handles such a model mismatch \textit{without the knowledge of the true distribution}, except for a bound on the KL divergence between the two.}
    \label{fig: distributions}
    %\vspace{-1.0em}
    \end{center}
\end{figure}

Our distributionally robust control algorithm can alternatively be viewed as an algorithm for automatic online tuning of the risk-sensitivity parameter in applying risk-sensitive control. Risk-sensitive optimal control has been shown to be effective and successful in many robotics applications \cite{medina2012risk, medina2012disagreement, bechtle2020curious, nishimura2020rssac}. However, in prior work the user has to specify a fixed risk-sensitivity parameter offline. This would require an extensive trial and error process until a desired robot behavior is observed. Furthermore, a risk-sensitivity parameter that works in a certain state can be infeasible in another state, as we will see in Section \ref{sec: algorithm}. Ideally, the risk-sensitivity should be adapted online depending on the situation to obtain a specifically desired robot behavior \cite{medina2012disagreement, nishimura2020rssac}, yet this is highly nontrivial as no simple general relationship is known between the risk-sensitivity parameter and the performance of the robot.
%while ideally it should be changed online depending on the situation to obtain a specifically desired robot behavior \cite{medina2012disagreement, nishimura2020rssac}. %the need for manually specifying the risk-sensitivity parameter has limited its applicability. % it would be useful to add sth in here on why it is hard to pick risk-sensitivity parameter, why we may need to change it throughout the trajectory.... 
Our algorithm addresses this challenge as a secondary contribution. %we demonstrate that eliminates such a requirement.
Due to the fundamental equivalence between distributionally robust control and risk-sensitive control, it serves as a nonlinear risk-sensitive control that can dynamically adjust the risk-sensitivity parameter depending on the state of the robot as well as the surrounding environment. %, given a KL divergence bound characterizing the distributional mismatch.

The rest of the paper is organized as follows. Section \ref{sec: related} reviews the related work in controls and robotics literature. Section \ref{sec: problem} summarizes the theoretical results originally presented in \cite{petersen2000minimax} %. As we will see, the theory naturally leads to a bilevel optimization problem that consists of risk-sensitive optimal control as the inner loop and one-dimensional optimization that serves as the outer loop. However, it is highly nontrivial to solve this problem in general cases. 
that connect distributionally robust control to risk-sensitive optimal control.
Section \ref{sec: algorithm} develops this theory into an algorithm that provides a locally optimal solution for general nonlinear systems with non-convex cost functions, which is a novel contribution of this paper. In Section \ref{sec: results}, we test our method in a collision avoidance scenario wherein the predictive distribution of pedestrian motion is erroneous. We further show its benefits as a risk-sensitive optimal controller that can automatically adjust its risk-sensitivity parameter in this section. The paper concludes in Section \ref{sec: conclusions} with potential future research directions.

%% file: related_work.tex
\subsection{Distributional Robustness and Risk-Sensitivity}
%We %briefly note that there exist various formulations to account for distributional robustness in optimal control. They share the common motivation to address distributional mismatch between the model and the true unknown system within an ambiguity set, a set of probability distributions in which the true distribution is contained. % references?
%Among all these notions of distributional robustness, it is the definition of the ambiguity set that varies across existing methods, which also leads to different optimization algorithms. Some methods require knowledge of the moments of the true distribution to construct the ambiguity set~ \cite{parys2016robust, samuelson2017data} while other methods define the ambiguity sets using divergence measures. Sinha et al. \cite{sinha2020formulazero} adopt the $\chi^2$-divergence as well as more general divergence measures in the space of probability distributions. The Wasserstein metric is employed in \cite{hakobyan2020wasserstein}. Similar to \cite{sinha2020formulazero}, we define the ambiguity set based on the KL divergence. However, our solution method relies on the transformation of the original problem into a series of nonlinear risk-sensitive optimal control problems that are solved online.
Distributionally robust control seeks to optimize control actions against a worst-case distribution within a given set of probability distributions, often called the ambiguity set \cite{parys2016robust, samuelson2017data}. There exist various formulations to account for distributional robustness in optimal control. Some works are concerned with minimizing the worst-case expectation of a cost objective \cite{samuelson2017data, sinha2020formulazero}, while others enforce risk-based or chance constraint satisfaction under a worst-case distribution \cite{hakobyan2020wasserstein, parys2016robust}. The present work belongs to the former class. Existing methods also differ in the formulation of the ambiguity set. Moment-based ambiguity sets require knowledge of moments of the ground-truth distribution up to a finite order \cite{parys2016robust, samuelson2017data}, which is often overly conservative \cite{hakobyan2020wasserstein}. Statistical distance-based ambiguity sets are also gaining attention. The authors of \cite{hakobyan2020wasserstein} use a Wasserstein metric to define the ambiguity set for motion planning with collision avoidance, but their MPC formulation is not suited for nonlinear systems. $\chi^2$-divergence and more general $\phi$-divergences (which KL divergence belongs to) are employed in \cite{sinha2020formulazero}, similar to the present work. However, the ambiguity set considered in \cite{sinha2020formulazero} is restricted to categorical distributions, while our work requires no assumption on the class of the ground-truth distributions. Furthermore, we make use of risk-sensitive optimal control to obtain planned robot trajectories with feedback, unlike sampling in their implementation.

Optimization of the entropic risk measure has been an active research topic in economics and controls literature since 1970s \cite{jacobson1973optimal, whittle1981risk, whittle2002risk, exarchos2016game}. The concept of risk-sensitive optimal control has been successfully applied to robotics in various domains, including haptic assistance \cite{medina2012risk, medina2012disagreement}, model-based reinforcement learning (RL) \cite{bechtle2020curious}, and safe robot navigation \cite{nishimura2020rssac, wang2020game}, to name a few. In all these works, the risk-sensitivity parameter is introduced as a user-specified constant, and is found to significantly affect the behavior of the robot. For instance, our prior work on safe robot navigation in human crowds \cite{nishimura2020rssac} reveals that a robot with higher risk-sensitivity tends to yield more to oncoming human pedestrians. However, how to find a desirable risk-sensitivity parameter still remains an open research question; in the robot navigation problem, the robot simply freezes if it is too risk-sensitive when the scene is crowded. As the authors of \cite{medina2012disagreement} point out, the robot should adapt its risk-sensitivity level depending on the situation, yet there still does not exist an effective algorithmic framework to automate it due to the issues discussed in Section \ref{sec: intro}. In this work, we provide such an algorithm for nonlinear, non-Gaussian stochastic systems. As mentioned earlier, our approach is built on previously-established theoretical results that link risk-sensitive and distributionally robust control \cite{petersen2000minimax}.

\subsection{Approximate Methods for Optimal Feedback Control}
\label{sec: related_work_approximate_oc}
The theory of optimal control lets us derive an optimal feedback control law via dynamic programming (DP) \cite{bertsekas1976dynamic}. For linear systems with additive Gaussian white noise and quadratic cost functions, the exact DP solution is tractable and is known as Linear-Quadratic-Gaussian (LQG) \cite{astrom1970introduction} or Linear-Exponential-Quadratic-Gaussian (LEQG) \cite{whittle1981risk}. They are different in that LQG optimizes the expected cost while LEQG optimizes the entropic risk measure, although both DP recursions are quite similar.

However, solving general optimal control problems for nonlinear systems remains a challenge due to lack of analytical tractability. Hence, approximate local optimization methods have been developed, including Differential Dynamic Programming (DDP) \cite{jacobson1970differential}, iterative Linear-Quadratic Regulator (iLQR) \cite{li2004iterative}, and iterative Linear-Quadratic-Gaussian (iLQG) \cite{todorov2005generalized, tassa2012synthesis}. While both DDP and iLQR are designed for deterministic systems with quadratic cost functions, iLQG can locally optimize the expected cost objective for Gaussian stochastic systems with non-convex cost functions. Similarly, the iterative Linear-Exponential-Quadratic-Gaussian (iLEQG) has been recently proposed to locally optimize the entropic risk for Gaussian systems with non-convex costs \cite{farshidian2015risk, wang2020game, roulet2020convergence}. 
Note however that they are not designed to be robust to model mismatches that we consider in this paper. In fact, it is known that even LQG does not possess guaranteed robustness \cite{doyle1978guaranteed}.

%Our bilevel optimization uses iLEQG for inner loop minimization, but note that unlike prior work \cite{farshidian2015risk, roulet2020convergence, wang2020game} our formulation does not require the Gaussian assumption on the true distribution (see Figure \ref{fig: distributions}).

%% file: problem_statement.tex
\subsection{Distributionally Robust Optimal Control}
\label{sec: problem_robust_control}

Consider the following stochastic nonlinear system:
\begin{align}
    x_{k+1} = f(x_k, u_k) + g(x_k, u_k)w_k,
    \label{eq: reference_system}
\end{align}
where $x_k \in \mathbb{R}^n$ denotes the state, $u_k \in \mathbb{R}^m$ the control, and $w_k \in \mathbb{R}^r$ the noise input to the system at time $k$. For some finite horizon $N$, let $w_{0:N} \triangleq (w_0,\dots, w_N)$ denote the joint noise vector with probability distribution $q(w_{0:N})$. This distribution is assumed to be a known Gaussian white noise process, i.e. $w_i$ is independent of $w_j$ for all $ i \neq j$, and we call \eqref{eq: reference_system} the reference system. Ideally, we would like the model distribution $q$ to perfectly characterize the noise in the dynamical system. However, in reality the noise may come from a different, more complex distribution which we may not know exactly. %Let us denote by $\bar{w}_{0:N} \triangleq (\bar{w}_0, \dots, \bar{w}_N)$ with probability distribution $q(\bar{w}_{0:N})$ a perturbed noise vector. 
Let $\bar{w}_{0:N} \triangleq (\bar{w}_0, \dots, \bar{w}_N)$ denote a perturbed noise vector that is distributed according to $p(\bar{w}_{0:N})$.
%We define the following perturbed system:
We define the following perturbed system that characterizes the true but unknown dynamics:
\begin{align}
    x_{k+1} = f(x_k, u_k) + g(x_k, u_k) \bar{w}_k.
    \label{eq: perturbed_system}
\end{align}
Note that we make no assumptions on Gaussianity or whiteness of $p$. One could also attribute it to potentially unmodeled dynamics.
The true, unknown probability distribution $p$ is contained in the set $\mathcal{P}$ of all probability distributions on the support $\mathbb{R}^{r(N+1)}$. We assume that $p$ is not ``too different" from $q$. This is encoded as the following constraint on the KL divergence between $p$ and $q$:
\begin{align}
    \mathbb{D}_{\mathrm{KL}}(p \Vert q) \leq d,
    \label{eq: relative_entropy_constraint}
\end{align}
where $\mathbb{D}_{\mathrm{KL}}(\cdot \Vert \cdot)$ is the KL divergence and $d > 0$ is a given constant. Note that $\mathbb{D}_{\mathrm{KL}}(p \Vert q) \geq 0$ always holds, with equality if and only if $p \equiv q$. The set of all possible probability distributions $p \in \mathcal{P}$ satisfying \eqref{eq: relative_entropy_constraint} is denoted by $\Xi$, which we define as our ambiguity set. This set is a convex subset of $\mathcal{P}$ for a fixed $q$ (Lemma 1.4.3, \cite{dupuis2011weak}). %Note that $q$ can be an arbitrarily complex distribution as long as the KL divergence constraint \eqref{eq: relative_entropy_constraint} is met.

We are interested in controlling the perturbed system \eqref{eq: perturbed_system} with a state feedback controller of the form $u_k = \mathcal{K}(k, x_k)$.
%\begin{align}
%    u_k = \mathcal{K}(k, x_k).
%    \label{eq: admissible_controller}
%\end{align}
The operator $\mathcal{K}(k, \cdot)$ defines a mapping from $\mathbb{R}^n$ into $\mathbb{R}^m$. The class of all such controllers is denoted $\Lambda$.

%The objective function considered in this paper is a cost function given by
The cost function considered in this paper is given by
\begin{align}
    J(x_{0:N+1}, u_{0:N}) \triangleq \sum_{k=0}^N c(k, x_k, u_k) + h(x_{N+1}),
    \label{eq: cost_functional}
\end{align}
where $c$ is the stage cost function and $h$ is the terminal cost.
We assume that the above objective satisfies the following non-negativity assumption.
\begin{assumption}[Assumption 3.1, \cite{petersen2000minimax}]
The functions $h(\cdot)$ and $c(k, \cdot, \cdot)$ satisfy $h(x) \geq 0$ and $c(k, x, u) \geq 0$ for all $k \in \{0, \dots, N\}$, $x \in \mathbb{R}^n$, and $u \in \mathbb{R}^m$.
\label{assumption: non_negative_costs}
\end{assumption}

Under the dynamics model \eqref{eq: perturbed_system}, the cost model \eqref{eq: cost_functional}, and the KL divergence constraint \eqref{eq: relative_entropy_constraint} on $p$, we are interested in finding an admissible controller $\mathcal{K} \in \Lambda$ that minimizes the worst-case expected value of the cost objective \eqref{eq: cost_functional}. In other words, we are concerned with the following distributionally robust optimal control problem:
\begin{align}
    \inf_{\mathcal{K} \in \Lambda} \sup_{
    p\in \Xi} \mathbb{E}_{p}\left[J(x_{0:N+1}, u_{0:N})\right],
    \label{eq: minimax_control}
\end{align}
where $\mathbb{E}_{p}[\cdot]$ indicates that the expectation is taken with respect to the true, unknown distribution $p$. In this formulation, the robustness arises from the ability of the controller to plan against a worst-case distribution $p$ %that is chosen adversarially from the ambiguity set $\Xi$.
in the ambiguity set $\Xi$.
\begin{remark}
If the KL divergence bound $d$ is zero, then $p \equiv q$ is necessary. In this degenerate case, \eqref{eq: minimax_control} reduces to the standard stochastic optimal control problem:
\begin{align}
    \inf_{\mathcal{K} \in \Lambda} \mathbb{E}_{q}\left[J(x_{0:N+1}, u_{0:N})\right].
\end{align}
\label{remark: case_d_is_zero}
\end{remark}

\subsection{Equivalent Risk-Sensitive Optimal Control}
\label{sec: problem_risk_sensitive_oc}
Unfortunately, the distributionally robust optimal control problem \eqref{eq: minimax_control} is intractable as it involves maximization with respect to a probability distribution $p$. To circumvent this, \cite{petersen2000minimax} proves that problem \eqref{eq: minimax_control} is equivalent to a bilevel optimization problem involving risk-sensitive optimal control with respect to the model distribution $q$. We refer the reader to \cite{petersen2000minimax} for the derivation and only re-state the main results in this section for self-containedness. Before doing so, we impose an additional assumption on the worst-case expected cost.
\begin{assumption}[Assumption 3.2, \cite{petersen2000minimax}]
For any admissible controller $\mathcal{K} \in \Lambda$, the resulting closed-loop system satisfies
\begin{align}
    \sup_{p\in \mathcal{P}} \mathbb{E}_{p} \left[J(x_{0:N+1}, u_{0:N})\right] = \infty.
\end{align}
\label{assumption: inf_cost}
\end{assumption}
This assumption states that, without the KL divergence constraint, some adversarially-chosen noise could make the expected cost objective arbitrarily large in the worst case. It amounts to a controllability-type assumption with respect to the noise input and an observability-type assumption with respect to the cost objective \cite{petersen2000minimax}.

Under Assumptions \ref{assumption: non_negative_costs} and \ref{assumption: inf_cost}, the following theorem holds.
\begin{theorem}
\label{thm: equivalence}
Consider the stochastic systems \eqref{eq: reference_system}, \eqref{eq: perturbed_system} with the KL divergence constraint \eqref{eq: relative_entropy_constraint} and the cost model \eqref{eq: cost_functional}. Under Assumptions \ref{assumption: non_negative_costs} and \ref{assumption: inf_cost}, the following equivalence holds for the distributionally robust optimal control problem \eqref{eq: minimax_control}:
\begin{multline}
    \inf_{\mathcal{K} \in \Lambda} \sup_{p \in \Xi} \mathbb{E}_{p}\left[J(x_{0:N+1}, u_{0:N})\right] \\
    = \inf_{\tau \in \tilde{\Gamma}} \inf_{\mathcal{K} \in \Lambda} \tau \log \mathbb{E}_{q}\left[\exp\left(\frac{J(x_{0:N+1}, u_{0:N})}{\tau}\right)\right] + \tau d,
    \label{eq: equivalence}
\end{multline}
provided that the set 
\begin{align}
\tilde{\Gamma} \triangleq \left\{\tau > 0: \inf_{\mathcal{K} \in \Lambda} \tau \log \mathbb{E}_{q}\left[\exp(J/\tau)\right]\text{ is finite}\right\}
\end{align}
is non-empty.
\end{theorem}
\begin{proof}
See Theorems 3.1 and 3.2 in \cite{petersen2000minimax}.
\end{proof}

\begin{remark}
Notice that the first term in the right-hand side of \eqref{eq: equivalence} is the entropic risk measure $R_{q,\frac{1}{\tau}}(J)$, where the risk is computed with respect to the model distribution $q$ and $\tau > 0$ serves as the inverse of the risk-sensitivity parameter. Rewriting the equation in terms of the risk-sensitivity parameter $\theta = 1/{\tau} > 0$, we see that the right-hand side of \eqref{eq: equivalence} is equivalent to
\begin{align}
    \inf_{\theta \in \Gamma}\left(\inf_{\mathcal{K} \in \Lambda} R_{q,\theta}\left(J(x_{0:N+1}, u_{0:N})\right) + \frac{d}{\theta}\right),
    \label{eq: bilevel_optimization}
\end{align}
where $\Gamma \triangleq \left\{\theta > 0: \inf_{\mathcal{K} \in \Lambda} R_{q,\theta}(J) \text{ is finite}\right\}$.
The non-emptiness of $\Gamma$ (and equivalently, $\tilde{\Gamma}$) is satisfied if there exists some non-zero risk-sensitivity $\theta$ that gives a finite entropic risk value. This is almost always satisfied in practical situations where risk-sensitive optimal control can be applied, as otherwise the problem would be ill-formed.
Theorem \ref{thm: equivalence} shows that the original distributionally robust optimal control problem \eqref{eq: minimax_control} is mathematically equivalent to a bilevel optimization problem \eqref{eq: bilevel_optimization} involving risk-sensitive optimal control. Note that the new problem does not involve any optimization with respect to the unknown distribution $p$.
\end{remark}

%% file: algorithm.tex
Even though the mathematical equivalence shown in \cite{petersen2000minimax} and summarized in Section \ref{sec: problem_risk_sensitive_oc} is general, it does not immediately lead to a tractable method to efficiently solve \eqref{eq: bilevel_optimization} for general nonlinear systems. There are two major challenges to be addressed. First, exact optimization of the entropic risk with a state feedback control law is intractable, except for linear systems with quadratic costs. Second, the optimal risk-sensitivity parameter has to be searched efficiently over the feasible space $\Gamma$, which not only is unknown but also varies dependent on the initial state $x_0$. A novel contribution of this paper is a tractable algorithm that approximately solves both of the problems for general nonlinear systems with non-convex cost functions. In what follows, we detail how we solve both the inner and the outer loop of \eqref{eq: bilevel_optimization} to develop a distributionally-robust, risk-sensitive MPC. 

\subsection{Iterative Linear-Exponential-Quadratic-Gaussian}
\label{sec: algorithm_ileqg}
Let us first consider the inner minimization of \eqref{eq: bilevel_optimization}:
\begin{align}
    \inf_{\mathcal{K} \in \Lambda} R_{q,\theta}\left(J(x_{0:N+1}, u_{0:N})\right),
    \label{eq: inner_optimization}
\end{align}
where we omitted the extra term $d/\theta$ as it is constant with respect to the controller $\mathcal{K}$. This amounts to solving a risk-sensitive optimal control problem for a nonlinear Gaussian system. Recently, a computationally-efficient, local optimization method called iterative Linear-Exponential-Quadratic-Gaussian (iLEQG) has been proposed for both continuous-time systems \cite{farshidian2015risk} and the discrete-time counterpart \cite{roulet2020convergence,wang2020game}. %iLEQG 
Both versions locally optimize the entropic risk measure with respect to a receding horizon, affine feedback control law for general nonlinear systems with non-convex costs.

%We adopt a variant of the discrete-time iLEQG algorithm \cite{wang2020game} to solve \eqref{eq: inner_optimization} for a locally optimal controller $\mathcal{K}$ of the following form:
%\begin{align}
%    \mathcal{K}(k, x_k) = L_k (x_k - \bar{x}_{k}) + l_k,
%\end{align}
%where $L_k \in \mathbb{R}^{m\times n}$ denotes the feedback gain matrix, $l_k \in\mathbb{R}^m$ the nominal control term, and $\bar{x}_{0:N+1}$ the nominal state trajectory obtained by applying $l_{0:N}$ to the noiseless dynamics $\bar{x}_{k+1} = f(\bar{x}_k, l_k)$.
%\begin{align}
%    \bar{x}_{k+1} = f(\bar{x}_k, l_k).
%\end{align}
%Note that the Gaussian assumption in this work is merely for the reference distribution $p$ and we still allow the actual distribution $q$ to be arbitrary within the given KL divergence bound $d$.
We adopt a variant of the discrete-time iLEQG algorithm \cite{wang2020game} to obtain a locally optimal solution to \eqref{eq: inner_optimization}. In what follows, we assume that the noise coefficient function $g(x_k, u_k)$ in \eqref{eq: reference_system} is the identity mapping for simplicity, but it is straightforward to handle nonlinear functions in a similar manner as discussed in \cite{todorov2005generalized}. The algorithm starts by applying a given nominal control sequence $l_{0:N}$ to the noiseless dynamics to obtain the corresponding nominal state trajectory $\bar{x}_{0:N+1}$. In each iteration, the algorithm maintains and updates a locally optimal controller $\mathcal{K}$ of the form:
\begin{align}
    \mathcal{K}(k, x_k) = L_k (x_k - \bar{x}_{k}) + l_k,
\end{align}
where $L_k \in \mathbb{R}^{m\times n}$ denotes the feedback gain matrix. The $i$-th iteration of our iLEQG implementation consists of the following four steps:
%In what follows, we assume that the noise coefficient function $g(x_k, u_k)$ in \eqref{eq: reference_system} is the identity mapping for simplicity, but it is straightforward to handle nonlinear functions in a similar manner as discussed in \cite{todorov2005generalized}. The algorithm starts with a given nominal trajectory $\{l_{0:N}, \bar{x}_{0:N+1}\}$ with the gain matrices $L_{0:N}$ initialized as 0. It then iteratively updates the nominal trajectory as well as the feedback gain matrices until convergence, while making locally-linear and locally-quadratic approximations to the dynamics and the costs, respectively. The $i$-th iteration of our iLEQG implementation consists of the following four steps:
\begin{enumerate}[label=\textbf{\arabic*})]
    \item \textbf{Local Approximation:} Given the nominal trajectory $\{l^{(i)}_{0:N}, \bar{x}^{(i)}_{0:N+1}\}$, we compute the following linear approximation of the dynamics as well as the quadratic approximation of the cost functions:
    \begin{align}
        A_k &= D_x f(\bar{x}^{(i)}_k, l^{(i)}_k), B_k = D_u f(\bar{x}^{(i)}_k, l^{(i)}_k) \\
        q_k &= c(k, \bar{x}^{(i)}_k, l^{(i)}_k) \\ 
        \mathbf{q}_k &= D_{x} c(k, \bar{x}^{(i)}_k, l^{(i)}_k), Q_k = D_{xx} c(k, \bar{x}^{(i)}_k, l^{(i)}_k) \\
        \mathbf{r}_k &= D_{u} c(k, \bar{x}^{(i)}_k, l^{(i)}_k) \\
        R_k &= D_{uu} c(k, \bar{x}^{(i)}_k, l^{(i)}_k), P_k = D_{ux} c(k, \bar{x}^{(i)}_k, l^{(i)}_k)
    \end{align}
    for $k = 0$ to $N$, where $D$ is the differentiation operator. We also let
    $q_{N + 1} = h(\bar{x}^{(i)}_{N+1})$, $\mathbf{q}_{N + 1} = D_x h(\bar{x}^{(i)}_{N+1})$, and $Q_{N + 1} = D_{xx} h(\bar{x}^{(i)}_{N+1})$.
    
    \item \textbf{Backward Pass:} We perform approximate DP using the current feedback gain matrices $L^{(i)}_{0:N}$ as well as the approximated model obtained in the previous step. Suppose that the noise vector $w_k$ is Gaussian-distributed according to $\mathcal{N}(0, W_k)$ with $W_k \succ 0$. Let $s_{N+1} \triangleq q_{N+1}$, $\mathbf{s}_{N+1} \triangleq \mathbf{q}_{N+1}$, and $S_{N+1} \triangleq Q_{N+1}$. Given these terminal conditions, we recursively compute the following quantities:
    \begin{align}
        M_k &= W_k^{-1} - \theta S_{k+1} \\
        \mathbf{g}_k &= \mathbf{r}_k + B_k^\mathrm{T}(I + \theta S_{k+1} M_k^{-1}) \mathbf{s}_{k+1} \\
        G_k &= P_k + B_k^\mathrm{T}(I + \theta S_{k+1} M_k^{-1})S_{k+1}A_k \\
        H_k &= R_k + B_k^\mathrm{T}(I + \theta S_{k+1} M_k^{-1})S_{k+1}B_k,
    \end{align}
    and
    \begin{multline}
        s_k = q_k + s_{k+1} - \frac{1}{2\theta} \log\det(I - \theta W_k S_{k+1}) \\
        + \frac{\theta}{2}\mathbf{s}_{k+1}^{\mathrm{T}}M_k^{-1}\mathbf{s}_{k+1} + \frac{1}{2} l^{(i) \mathrm{T}}_k H_k l^{(i)}_k + l^{(i) \mathrm{T}}_k \mathbf{g}_k
    \end{multline}
    \vspace{-1.5em}
    \begin{multline}
        \mathbf{s}_k = \mathbf{q}_k + A_k^{\mathrm{T}}(I + \theta S_{k+1}M_k^{-1})\mathbf{s}_{k+1} \\ + L^{(i)\mathrm{T}}_k H_k l^{(i)}_k
        + L_k^{(i)\mathrm{T}} \mathbf{g}_k + G_k^{\mathrm{T}} l^{(i)}_k
    \end{multline}
    \vspace{-1.5em}
    \begin{multline}
        S_{k} = Q_k +  A_k^{\mathrm{T}}(I + \theta S_{k+1}M_k^{-1})S_{k+1}A_k \\
        + L_k^{(i)\mathrm{T}}H_k L_k^{(i)} + L_k^{(i)\mathrm{T}}G_k + G_k^{\mathrm{T}}L_k^{(i)},
    \end{multline}
    from $k = N$ down to $0$. Note that $M_k \succ 0$ is necessary so it is invertible, which may not hold if $\theta$ is too large. This is called ``neurotic breakdown," when the optimizer is so pessimistic that the cost-to-go approximation becomes infinity \cite{whittle2002risk}. Otherwise, the approximated cost-to-go for this optimal control (under the controller $\{L^{(i)}_{0:N}, l^{(i)}_{0:N}\}$) is given by $s_0$.
    
    \item \textbf{Regularization and Control Computation:} Having derived the DP solution, we compute new control gains $L^{(i+1)}_{0:N}$ and offset updates $dl_{0:N}$ as follows:
    \begin{align}
        L^{(i+1)}_k &= -(H_k + \mu I)^{-1} G_k \\
        dl_k &= -(H_k + \mu I)^{-1} \mathbf{g}_k ,
    \end{align}
    where $\mu \geq 0$ is a regularization parameter to prevent $(H_k + \mu I)$ from having negative eigenvalues. We adaptively change $\mu$ across multiple iterations as suggested in \cite{tassa2012synthesis}, so the algorithm enjoys fast convergence near a local minimum while ensuring the positive-definiteness of $(H_k + \mu I)$ at all times.
    
    \item \textbf{Line Search for Ensuring Convergence:}
    It is known that the update could lead to increased cost or even divergence if a new trajectory strays too far from the region where the local approximation is valid \cite{tassa2012synthesis}. Thus, the new nominal control trajectory $l^{(i+1)}_{0:N}$ is computed by backtracking line search with line search parameter $\epsilon$. Initially, $\epsilon = 1$ and we derive a new candidate nominal trajectory as follows:
    \begin{align}
        \hat{l}_k &= L_k^{(i+1)}(\hat{x}_k - \bar{x}^{(i)}_k) + l^{(i)}_k + \epsilon dl_k \\
        \hat{x}_{k+1} &= f(\hat{x}_k, \hat{l}_k).
    \end{align}
    If this candidate trajectory $\{\hat{l}_{0:N}, \hat{x}_{0:N+1}\}$ results in a lower cost-to-go than the current nominal trajectory, then the candidate trajectory is accepted and returned as $\{l^{(i+1)}_{0:N}, \bar{x}^{(i+1)}_{0:N+1}\}$. Otherwise, the trajectory is rejected and re-derived with $\epsilon \leftarrow \epsilon/2$ until it is accepted. More details on this line search can be found in \cite{van2012motion}.
\end{enumerate}

The above procedure is iterated until the nominal control $l_k$ does not change beyond some threshold in a norm. Once converged, the algorithm returns the nominal trajectory $\{l_{0:N}, \bar{x}_{0:N+1}\}$ as well as the feedback gains $L_{0:N}$ and the approximate cost-to-go $s_0$.

\subsection{Cross-Entropy Method}
\label{sec: algorithm_ce}
Having implemented the iLEQG algorithm as a local approximation method\footnote{Although the solutions obtained by iLEQG are local optima and different for varied risk-sensitivity $\theta$, we have empirically observed that they are not changing drastically as a function of $\theta$. This indicates that iLEQG is a sensible choice for the inner-loop optimization problem, since the solutions do not hop around local optima that are too different as we vary $\theta$.} for the inner-loop optimization of \eqref{eq: bilevel_optimization}, it remains to solve the outer-loop optimization for the optimal risk-sensitivity parameter $\theta^*$. This is a one-dimensional optimization problem in which the function evaluation is done by solving the corresponding risk-sensitive optimal control \eqref{eq: inner_optimization}. In this work we choose to adapt the cross entropy method \cite{rubinstein2013cross, kochenderfer2019algorithms} to derive the approximately optimal value for $\theta^*$. This method is favorable for online optimization due to its any-time nature and high parallelizability of the Monte Carlo sampling. As a byproduct, it is also effective in approximately finding the maximum feasible $\theta \in \Gamma$ within a few iterations, which is detailed later in this section. We note however that it is possible to use other methods for the outer-loop optimization as well. The cross entropy method is a stochastic method that maintains an explicit probability distribution over the design space. At each step, a set of $m_{\text{s}}$ Monte Carlo samples is drawn from the distribution, out of which a subset of $m_{\text{e}}$ ``elite samples" that achieve the best performance is retained. The parameters of the distribution is then updated according to the maximum likelihood estimate on the elite samples. The algorithm stops after a desired number of steps $M$. %The cross entropy method is favorable for online optimization due to its any-time nature and high parallelizability of the Monte Carlo sampling.

In our implementation we model the distribution\footnote{Note that this distribution is defined in the space of $\theta$ and has nothing to do with $p$ or $q$ that define stochastic noise in the dynamical system.} as univariate Gaussian $\mathcal{N}(\mu, \sigma^2)$. A remaining issue is that the iLEQG may return the cost-to-go of infinity if a sampled $\theta$ is too large, due to neurotic breakdown. Since our search space is limited to $\Gamma$ where $\theta$ yields a finite cost-to-go, we have to ensure that each iteration has enough samples in $\Gamma$. %Furthermore, the region of such valid $\theta$ may vary as the iLEQG is initialized with different nominal trajectories, since it is a local approximation method.

To address this problem, we augment the cross entropy method with rejection and re-sampling. Out of the $m_\text{s}$ samples drawn from the univariate Gaussian, we first discard all non-positive samples. For each of the remaining samples, we evaluate the objective \eqref{eq: bilevel_optimization} by a call to iLEQG, and then count the number of samples that obtained a finite cost-to-go. Let $m_{\text{v}}$ be the number of such valid samples. If $m_{\text{v}} \geq \max(m_{\text{e}}, m_{\text{s}}/2)$, we proceed and fit the distribution. Otherwise, we redo the sampling procedure as there are not sufficiently many valid samples to choose the elites from. 

In practice, re-sampling is not likely to occur after the first iteration of the cross entropy method. At the same time, we empirically found that the first iteration has a risk of re-sampling multiple times, hence degrading the efficiency. We therefore also perform an adaptive initialization of the Gaussian parameters $\mu_{\text{init}}$ and $\sigma_{\text{init}}$ in the first iteration as follows. If the first iteration with $\mathcal{N}(\mu_{\text{init}}, \sigma^2_{\text{init}})$ results in re-sampling, we not only re-sample but also divide $\mu_{\text{init}}$ and $\sigma_{\text{init}}$ by half. If all of the $m_{\text{s}}$ samples are valid, on the other hand, we accept them but double $\mu_{\text{init}}$ and $\sigma_{\text{init}}$, since it implies that the initial set of samples is not wide-spread enough to cover the whole feasible set $\Gamma$. The parameters $\mu_{\text{init}}$ and $\sigma_{\text{init}}$ are stored internally in the cross entropy solver and carried over to the next call to the algorithm. We have empirically found that this adaptive initialization is also useful for approximately finding the maximum feasible $\theta$, which we exploited in a comparative study in Section \ref{sec: optimal_risk_sensitivity}.

\subsection{RAT iLQR as MPC}
\label{sec: algorithm_dr_ileqg}
 We name the proposed bilevel optimization algorithm RAT iLQR. The pseudo-code is given in Algorithm \ref{algo: dr_ileqg}. At run time, it is executed as an MPC in a receding-horizon fashion; the control is re-computed after executing the first control input $u_0 = l_0$ and transitioning to a new state. A previously-computed control trajectory $l_{0:N}$ is reused for the initial nominal control trajectory at the next time step to warm-start the computation.
 
\begin{figure}
\begin{minipage}{\dimexpr\linewidth}
\begin{algorithm}[H]
	\caption{RAT iLQR Algorithm}\label{algo: dr_ileqg}
	\begin{algorithmic}[1]
		\INPUT Initial state $x_0$, controls $l_{0:N}$, $L_{0:N}$ (can be zero), KL divergence bound $d$
		\OUTPUT New nominal trajectory $\{l_{0:N}, \bar{x}_{0:N+1}\}$, control gains $L_{0:N}$, risk-sensitivity parameter $\theta^*$
		\State Compute initial nominal trajectory $\bar{x}_{0:N+1}$ using $l_{0:N}$
		\State $i \leftarrow 1$ 
		\While{$i \leq M$}   /* outer-loop optimization */
		\While{True}
		\If{$i = 1$}
		\State $\theta_{\text{sampled}} \leftarrow drawSamples(m_{\text{s}}, \mu_{\text{init}}, \sigma_{\text{init}})$
		\Else
		\State $\theta_{\text{sampled}} \leftarrow drawSamples(m_{\text{s}}, \mu, \sigma)$
		\EndIf
		\State array $r$ \Comment{Empty array of size $m_{\text{s}}$}
		\For{$j \leftarrow 1:m_{\text{s}}$} /* inner-loop optimization */
		\State Solve iLEQG with $\{l_{0:N}, \bar{x}_{0:N+1}, \theta_{\text{sampled}}[j]\}$
		\State Obtain approximate cost-to-go $s_0$
		\State $r[j] \leftarrow s_0 + d/\theta_{\text{sampled}}[j]$
		\EndFor
		\State $m_{\text{v}} \leftarrow countValidSamples(\theta_{\text{sampled}}, r)$
		\If{$i = 1$ and $m_{\text{v}} < \max(m_{\text{e}}, m_{\text{s}}/2)$}
		\State $\mu_{\text{init}} \leftarrow \mu_{\text{init}}/2, \sigma_{\text{init}} \leftarrow \sigma_{\text{init}}/2$
		\ElsIf{$i = 1$ and $m_{\text{v}} = m_{\text{s}}$}
		\State $\mu_{\text{init}} \leftarrow 2\mu_{\text{init}}, \sigma_{\text{init}} \leftarrow 2\sigma_{\text{init}}$
		\State \textbf{break}
		\ElsIf{$m_{\text{v}} \geq \max(m_{\text{e}}, m_{\text{s}}/2)$}
		\State \textbf{break}
		\EndIf
		\EndWhile
		\State $\theta_{\text{elite}} \leftarrow selectElite(m_{\text{e}}, \theta_{\text{sampled}}, r)$
		\State $\{\mu, \sigma\} \leftarrow fitGaussian(\theta_{\text{elite}})$
		\State $i \leftarrow i + 1$
		\EndWhile
		\State $\theta^* \leftarrow \mu$
		\State Solve iLEQG with $\{l_{0:N}, \bar{x}_{0:N+1}, \theta^*\}$
		%\State Obtain new $\{l_{0:N}, \bar{x}_{0:N+1}\}$ and $L_{0:N}$
		\State \textbf{return} new $\{l_{0:N}, \bar{x}_{0:N+1}\}$ with $L_{0:N}$ and $\theta^*$
	\end{algorithmic}
\end{algorithm}
\end{minipage}
%\vspace{-3mm}
%\vspace{-1.5em}
\end{figure}

%% file: results.tex
This section presents qualitative and quantitative results of the simulation study that we conducted to show the effectiveness of the RAT iLQR algorithm. We provide the problem setup as well as implementation details in Section \ref{sec: collision_avoidance_problem}. The goals of this study are two-fold. First, we demonstrate that the robot controlled by RAT iLQR can successfully accomplish its task under the presence of stochastic disturbance, without access to the ground-truth distribution but the knowledge of the KL divergence bound. This is presented in Section \ref{sec: performance_comparison} with comparisons to (non-robust) iLQG and a model-based MPC with sampling from the true generative model. Second, Section \ref{sec: optimal_risk_sensitivity} focuses on the nonlinear risk-sensitive optimal control aspect of RAT iLQR to show its value as an algorithm that can optimally adjust the risk-sensitivity parameter online, which itself is a novel contribution.

\subsection{Problem Setup}
\label{sec: collision_avoidance_problem}
We consider a dynamic collision avoidance problem where a unicycle robot has to avoid a pedestrian in a collision course as illustrated in Figure \ref{fig: environment}. Collision avoidance problems are often modeled by stochastic optimal control in the autonomous systems literature \cite{kusumoto2019informed} and the human-robot interaction literature \cite{schmerling2018multimodal, ivanovic2020mats}, including our prior work \cite{nishimura2020rssac}.

The state of the robot is defined by $(r_x, r_y, v, \theta) \in \mathbb{R}^4$,
where $(r_x, r_y)$ [m] denotes the position, $v$ [m/s] the velocity, and $\theta$ [rad] the heading angle. The robot's control input is $u = (a, b) \in \mathbb{R}^2$, where $a$ [m/s$^2$] is the acceleration and $b$ [rad/s] is the angular velocity. The pedestrian is modeled as a single integrator, whose position is given by $(p_x, p_y)$ [m]. We assume that the position of the pedestrian is known to the robot through onboard sensing. We also employ a constant nominal velocity model $(u_{x}, u_{y}) = (0.0, 1.0)$ [m/s] for the pedestrian. The joint system $x \in \mathbb{R}^6$ consists of the state of the robot and the position of the pedestrian. The dynamics are propagated by Euler integration with time interval $dt = 0.1$ [s] and additive noise $w_k$ to the joint state. The model distribution for $w_k$ is a zero-mean Gaussian $\mathcal{N}(0, W)$ with $W = \text{diag}([1e^{-10}, 1e^{-10}, 1e^{-3}, 1e^{-4}, 0.02, 0.02])\times dt$. This covariance matrix $W$ is chosen to encode our modeling assumption that the pedestrian motion is the main source of uncertainty in this joint system, and that the magnitude of the slip is almost negligible for the robot. The ground-truth distribution for the robot is the same Gaussian as in the model, but the pedestrian's distribution is a mixture of Gaussians that is independent of the robot's noise. Both the model and the true distributions for the pedestrian are illustrated in Figure \ref{fig: distributions}. Gaussian mixtures are favored by many recent papers in machine learning to account for multi-modality in human's decision making \cite{chai2019multipath, wang2020fast, salzmannivanovic2020trajectron++}. 

RAT iLQR requires an upper-bound on the KL divergence between the model and the true distribution. For the sake of this paper we assume that there is a separate module that provides an estimate. In this specific simulation study, we performed Monte Carlo integration with samples drawn from the true distribution offline. During the simulation, however, we did not reveal any information on the true distribution to RAT iLQR but the estimated KL value\footnote{One could test RAT iLQR under a distribution that has stronger multi-modality with a much larger KL bound than the one used in this simulation study, but it could introduce over-conservatism and lead to poor mean performance as the ambiguity set becomes too large. This is a common property of distributionally robust optimization \cite{kapteyn2018distributionally}.} of 32.02. This offline computation was possible due to our time-invariant assumption on the Gaussian mixture. If one is to use more realistic data-driven prediction instead, it is necessary to estimate the KL divergence online since the predictive distribution may change over time as the human-robot interaction evolves. Even though RAT iLQR works with time-varying KL bounds owing to its MPC formulation, we limit our attention to a static KL bound in this work as real-time computation of KL divergence can be challenging. Note that efficient and accurate estimation of information measures (including KL divergence) is still an active area of research in information theory and machine learning \cite{belghazi2018mine, iranzad2019estimation}, which is one of our future research directions.

The cost functions for this problem are given by
\begin{multline}
    c(k, x_k, u_k) = c_{\text{track}}(k, r_{x, k}, r_{y, k}, v_{k}, \theta_{k}) \\ + 
    c_{\text{coll}}(r_{x, k}, r_{y, k}, p_{x, k}, p_{y, k}) + c_{\text{ctrl}}(a_k, b_k),
\end{multline}
where $c_{\text{track}}$ denotes a quadratic cost that penalizes the deviation from a given target robot trajectory, $c_{\text{coll}}$ a collision penalty that incurs high cost when the robot is too close to the pedestrian, and $c_{\text{ctrl}}$ a small quadratic cost on the control input. Mirroring the formulation in \cite{wang2020game}, we used the following collision cost:
\begin{align}
    c_{\text{coll}} = \frac{10}{\left(0.2\sqrt{(r_{x} - p_x)^2 + (r_y - p_y)^2} + 0.9\right)^{10}}.
\end{align}

RAT iLQR was implemented in Julia and the Monte Carlo sampling of the cross entropy method was distributed across multiple CPU cores. Our implementation with $N = 19$, $M = 5$, $m_{\text{s}} = 10$, and $m_{\text{e}} = 3$ yielded the average computation time of 0.27 [s]. This is 2.7 times slower than real time, with $dt = 0.1$ [s] used to measure the real-time property. We expect to achieve improved efficiency by further parameter tuning as well as more careful parallelization.

\subsection{Comparison with Baseline MPC Algorithms}
\label{sec: performance_comparison}
\begin{figure}[t]
    \begin{center}
	\begin{tabular}{c}
		\begin{minipage}[t]{0.46\columnwidth}
			\centering
 			\scalebox{1.0}[1.0]{\includegraphics[trim=0 0 0 0,clip,width=1.0\columnwidth]{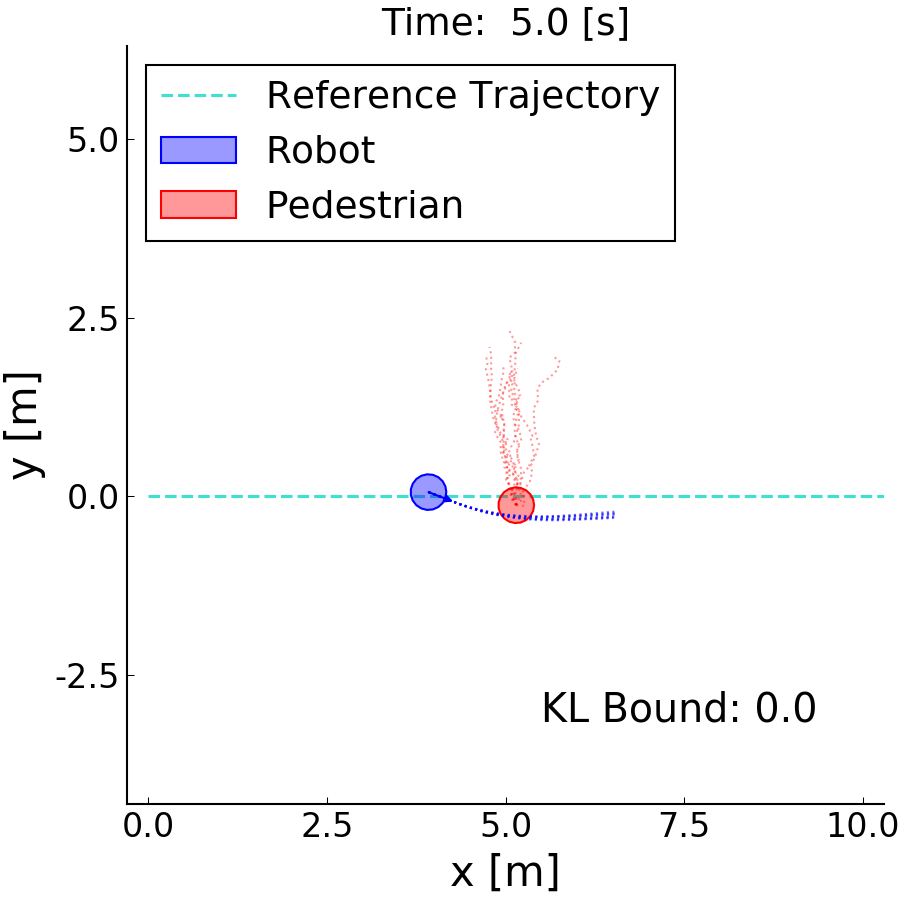}}
		\end{minipage}
		\begin{minipage}[t]{0.46\columnwidth}
			\centering
			\scalebox{1.0}[1.0]{\includegraphics[trim=0 0 0 0,clip,width=1.0\columnwidth]{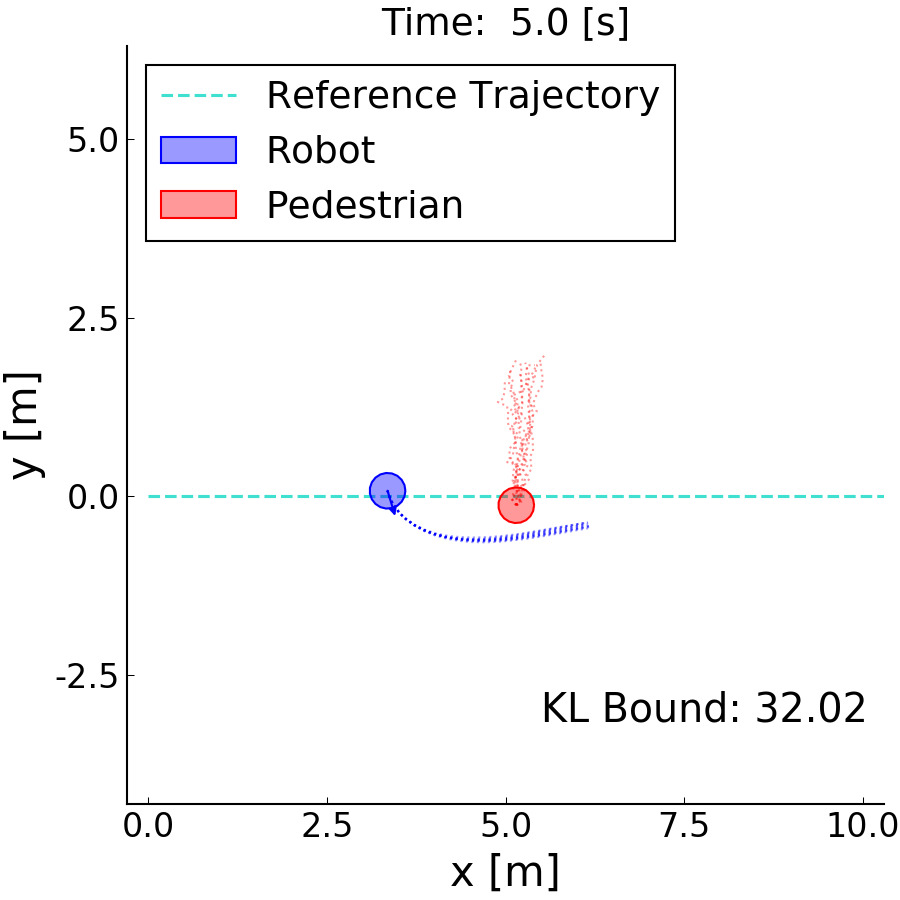}}
		\end{minipage}
	\end{tabular}
    \caption{A unicycle robot avoiding collision with a road-crossing pedestrian. (Left) When the KL bound is set to $d = 0$, RAT iLQR ignores this model error and reduces to iLQG. (Right) With the correct information on the KL, RAT iLQR is aware of the prediction error and optimally adjusts the risk-sensitivity parameter for iLEQG, planning a trajectory that stays farther away from the pedestrian. The figures are overlaid with predictions drawn from the model distribution and closed-loop motion plans of the robot. Note that the prediction for the pedestrian is erroneous since the actual pedestrian motion follows the Gaussian mixture distribution. The model distribution and the true Gaussian mixture are both illustrated in Figure \ref{fig: distributions}.}
    \label{fig: environment}
    %\vspace{-1.0em}
    \end{center}
\end{figure}

\begin{figure}[t]
    \centering
    \includegraphics[clip,width=0.80\columnwidth]{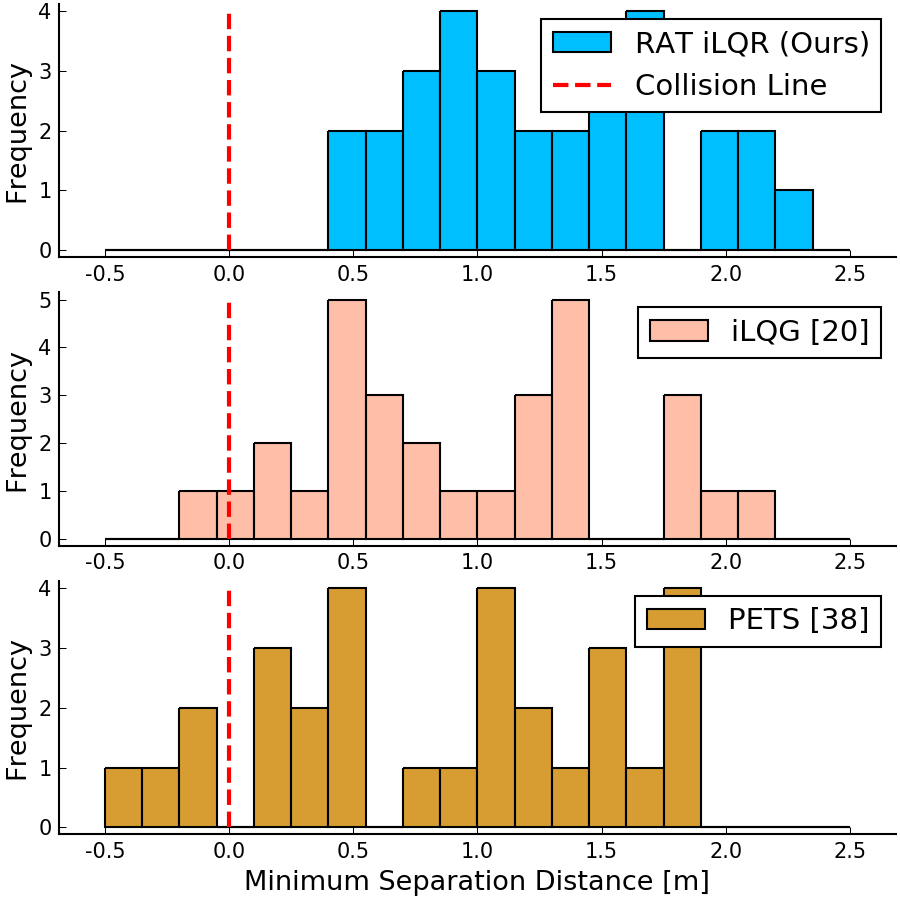}
    \caption{Histograms of the minimum separation distance between the robot and the pedestrian. A negative value indicates that a collision has occurred in that run. For each control algorithm, we performed 30 runs of the simulation with randomized pedestrian start positions. RAT iLQR consistently maintained a sufficient safety margin to avoid collision, while iLQG and PETS both failed. See Table \ref{table: histogram} for the summary statistics of these data.}
    \label{fig: histogram}
    %\vspace{-1.0em}
\end{figure}

We compared the performance of RAT iLQR against two baseline MPC algorithms, iLQG \cite{ todorov2005generalized} and PETS \cite{chua2018deep}. iLQG corresponds to RAT iLQR with the KL bound of $d = 0$, i.e. no distributional robustness is considered. Instead it is more computationally efficient than RAT iLQR, taking only 0.01 [s]. PETS is a state-of-the-art, model-based stochastic MPC algorithm with sampling and is originally proposed in a model-based reinforcement learning (RL) context. We chose PETS as our baseline since it also relies on the cross entropy method for online control optimization and is not limited to Gaussian distributions, similar to RAT iLQR. However, there are three major differences between PETS and RAT iLQR. First, PETS performs the cross entropy optimization directly in the high-dimensional control sequence space, which is far less sample efficient than RAT iLQR which uses the cross entropy method to only optimize the scalar risk-sensitivity parameter. Second, PETS does not consider feedback during planning as opposed to RAT iLQR. Third, PETS requires access to the exact ground-truth Gaussian mixture distribution to perform sampling, while RAT iLQR only relies on the KL divergence bound and the Gaussian distribution that we have modeled. We let PETS perform $M = 5$ iterations of the cross entropy optimization, each with $25$ samples for the control sequence coupled with $50$ samples for the joint state trajectory prediction, which resulted in the average computation time of 0.67 [s].

\begin{table}[t]
    \begin{center}
    \begin{tabular}{|l|c|c|}
    \hline
    Method & Min. Sep. Dist. [m] & Total Collision Count \\
    \hline
    RAT iLQR (Ours) & $\mathbf{1.26 \pm 0.51}$ & \textbf{0} \\
    \hline
    iLQG [20] & $0.94 \pm 0.62$ & 1 \\
    \hline
    PETS [38] & $0.87 \pm 0.69$ & 4 \\
    \hline
    \end{tabular}
    \caption{Statistics summarizing histogram plots presented in Figure \ref{fig: histogram}. RAT iLQR achieved the largest average value for the minimum separation distance with the smallest standard deviation, which contributed to safe robot navigation without a single collision. Note that PETS had multiple collisions despite its access to the true Gaussian mixture distribution.}
    \label{table: histogram}
    \end{center}
    %\vspace{-1.0em}
\end{table}

We performed 30 runs of the simulation for each algorithm, with randomized pedestrian start positions and stochastic transitions. To measure the performance, we computed the minimum separation distance between the robot and the pedestrian in each run, assuming that the both agents are circular with the same diameter. The histogram plots presented in Figure \ref{fig: histogram} clearly indicates the failure of iLQG and PETS as well as RAT iLQR's capability to maintain a sufficient safety margin for collision avoidance despite the distributional model mismatch. As summarized in Table \ref{table: histogram}, RAT iLQR achieved the largest minimum separation distance on average with the smallest standard deviation, which contributed to safe robot navigation. Note that even iLQG had one collision under this large model mismatch. Figure \ref{fig: environment} provides a qualitative explanation of this failure; the planned trajectories by iLQG tend to be much closer to the passing pedestrian than those by the risk-sensitive RAT iLQR. This difference is congruous with our earlier observations in prior work \cite{nishimura2020rssac} where risk-sensitivity is shown to affect the global behavior of the robot.

\subsection{Benefits of Risk-Sensitivity Parameter Optimization}
\label{sec: optimal_risk_sensitivity}
\begin{figure}[t]
    \centering
    \includegraphics[clip,width=0.8\columnwidth]{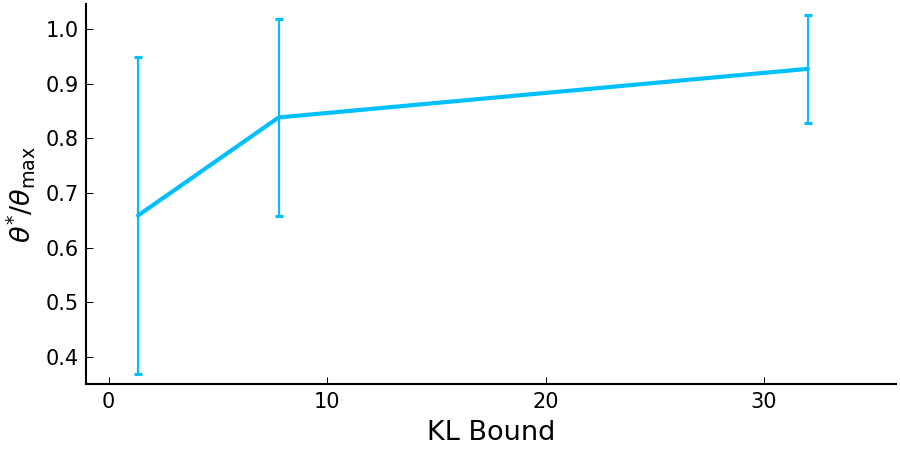}
    \caption{Time-averaged ratio of the optimal $\theta^*$ found by RAT iLQR to the maximum feasible $\theta$ before the neurotic breakdown occurs, plotted for three distinct KL divergence values. As the KL bound increases from 1.34 to 32.02, the ratio also consistently increased from 0.66 to 0.93. Note also that the standard deviation decreased from 0.29 to 0.10. This suggests that the robot becomes more risk-sensitive as the KL bound increases, and yet it does not choose the maximum $\theta$ value all the time.}
    \label{fig: theta_ratios}
    %\vspace{-1.0em}
\end{figure}

\begin{table}[t]
    \begin{center}
    \begin{tabular}{|l|c|c|}
    \hline
    \multicolumn{3}{|c|}{KL Bound: $d = 1.34$} \\
    \hline
    Method & Total Collision Count & Tracking Error [m] \\
    \hline
    RAT iLQR (Ours) & 0 & $\mathbf{0.22 \pm 0.29}$ \\
    \hline
    iLEQG with $\theta_{\max}$ & 0 & $0.36 \pm 0.45$\\
    \hline\hline
    \multicolumn{3}{|c|}{KL Bound: $d = 7.78$} \\
    \hline
    Method & Total Collision Count & Tracking Error [m] \\
    \hline
    RAT iLQR (Ours) & 0 & $\mathbf{0.25 \pm 0.35}$ \\
    \hline
    iLEQG with $\theta_{\max}$ & 0 & $0.38 \pm 0.52$ \\
    \hline\hline
    \multicolumn{3}{|c|}{KL Bound: $d = 32.02$} \\
    \hline
    Method & Total Collision Count & Tracking Error [m] \\
    \hline
    RAT iLQR (Ours) & 0 & $\mathbf{0.32 \pm 0.40}$ \\
    \hline
    iLEQG with $\theta_{\max}$ & 0 & $0.38 \pm 0.46$ \\
    \hline
    \end{tabular}
    \caption{Our comparative study between RAT iLQR with $\theta^*$ and iLEQG with $\theta_{\max}$ (i.e. maximum feasible risk-sensitivity) reveals that RAT iLQR's optimal choice of the risk-sensitivity parameter $\theta^{*}$ results in a more efficient robot navigation with smaller trajectory tracking errors, while still achieving collision avoidance under the model mismatch. With RAT iLQR, the average tracking error was reduced by $39\%$, $34\%$, and $16\%$, for 3 true distributions with different KL divergences of $1.34$, $7.78$, and $32.02$, respectively.}
    \label{table: ratilqr_vs_ileqg}
    \end{center}
    %\vspace{-1.0em}
\end{table}

 We also performed two additional sets of 30 simulations for RAT iLQR, with two different ground-truth distributions that are closer to the model distribution having the estimated KL divergence of 7.78 and 1.34, respectively. This is to study how different KL bounds affect the optimal choice of the risk-sensitivity parameter $\theta^*$. The results are shown in Figure \ref{fig: theta_ratios}. As the KL bound increases from 1.34 to 32.02, the ratio between the optimal $\theta^{*}$ found by RAT iLQR to the maximum feasible $\theta$ also increases. This matches our intuition that the larger the model mismatch, the more risk-sensitive the robot becomes. However, we also note that the robot does not saturate $\theta$ all the time even under the largest KL bound of 32.02. This raises a fundamental question on the benefits of RAT iLQR as a risk-sensitive optimal control algorithm: how favorable is RAT iLQR with optimal $\theta^{*}$ compared to iLEQG with the highest risk-sensitivity?

To answer this question, we performed a comparative study between RAT iLQR with $\theta^{*}$ and iLEQG with $\theta_{\max}$ (i.e. maximum feasible $\theta$ found during the cross entropy sampling of RAT iLQR) under the same simulation setup as before. The results are reported in Table \ref{table: ratilqr_vs_ileqg}. In terms of collision avoidance, both algorithms were equally safe with collision count of $0$. However, RAT iLQR achieved significantly more efficient robot navigation compared to iLEQG with $\theta_{\max}$, reducing the average tracking error by $39\%$, $34\%$, and $16\%$ for the KL values of $1.34$, $7.78$, and $32.02$, respectively. The efficiency and the safety of robot navigation are often in conflict in dynamic collision avoidance, and prior work \cite{nishimura2020rssac} struggles to find the right balance by manually tuning a fixed $\theta$. With RAT iLQR, such need for manual tuning is eliminated since the algorithm dynamically adjusts $\theta$ so it is the most desirable to handle the potential model mismatch specified by the KL bound.

%% file: conclusions.tex
In this work we propose RAT iLQR, a novel nonlinear MPC algorithm for distributionally robust control under a KL divergence bound. Our method is based on the mathematical equivalence between distributionally robust control and risk-sensitive optimal control. A locally optimal solution to the resulting bilevel optimization problem is derived with iLEQG and the cross entropy method. The simulation study shows that RAT iLQR successfully accounts for the distributional mismatch during collision avoidance. It also shows the effectiveness of dynamic adjustment of the risk-sensitivity parameter by RAT iLQR, which overcomes a limitation of conventional risk-sensitive optimal control methods. Future work will focus on accurate online estimation of the KL divergence from a stream of data. We are also interested in exploring applications of RAT iLQR, including control of learned dynamical systems and perception-aware control.

%% file: main.bbl
\begin{thebibliography}{10}
\providecommand{\url}[1]{#1}
\csname url@rmstyle\endcsname
\providecommand{\newblock}{\relax}
\providecommand{\bibinfo}[2]{#2}
\providecommand\BIBentrySTDinterwordspacing{\spaceskip=0pt\relax}
\providecommand\BIBentryALTinterwordstretchfactor{4}
\providecommand\BIBentryALTinterwordspacing{\spaceskip=\fontdimen2\font plus
\BIBentryALTinterwordstretchfactor\fontdimen3\font minus
  \fontdimen4\font\relax}
\providecommand\BIBforeignlanguage[2]{{%
\expandafter\ifx\csname l@#1\endcsname\relax
\typeout{** WARNING: IEEEtran.bst: No hyphenation pattern has been}%
\typeout{** loaded for the language `#1'. Using the pattern for}%
\typeout{** the default language instead.}%
\else
\language=\csname l@#1\endcsname
\fi
#2}}

\bibitem{petersen2000minimax}
I.~R. Petersen, M.~R. James, and P.~Dupuis, ``Minimax optimal control of
  stochastic uncertain systems with relative entropy constraints,'' \emph{IEEE
  Transactions on Automatic Control}, vol.~45, no.~3, pp. 398--412, 2000.

\bibitem{majumdar2020should}
A.~Majumdar and M.~Pavone, ``How should a robot assess risk? towards an
  axiomatic theory of risk in robotics,'' in \emph{Robotics Research}.\hskip
  1em plus 0.5em minus 0.4em\relax Springer, 2020, pp. 75--84.

\bibitem{whittle2002risk}
P.~Whittle, ``Risk sensitivity, a strangely pervasive concept,''
  \emph{Macroeconomic Dynamics}, vol.~6, no.~1, pp. 5--18, 2002.

\bibitem{medina2012risk}
J.~R. Medina, D.~Lee, and S.~Hirche, ``Risk-sensitive optimal feedback control
  for haptic assistance,'' in \emph{2012 IEEE international conference on
  robotics and automation}.\hskip 1em plus 0.5em minus 0.4em\relax IEEE, 2012,
  pp. 1025--1031.

\bibitem{medina2012disagreement}
J.~R. Medina, T.~Lorenz, D.~Lee, and S.~Hirche, ``Disagreement-aware physical
  assistance through risk-sensitive optimal feedback control,'' in \emph{2012
  IEEE/RSJ International Conference on Intelligent Robots and Systems}.\hskip
  1em plus 0.5em minus 0.4em\relax IEEE, 2012, pp. 3639--3645.

\bibitem{bechtle2020curious}
S.~Bechtle, Y.~Lin, A.~Rai, L.~Righetti, and F.~Meier, ``Curious ilqr:
  Resolving uncertainty in model-based rl,'' in \emph{Conference on Robot
  Learning}, 2020, pp. 162--171.

\bibitem{nishimura2020rssac}
H.~Nishimura, B.~Ivanovic, A.~Gaidon, M.~Pavone, and M.~Schwager,
  ``Risk-sensitive sequential action control with multi-modal human trajectory
  forecasting for safe crowd-robot interaction,'' in \emph{2020 IEEE/RSJ
  International Conference on Intelligent Robots and Systems (IROS)}.\hskip 1em
  plus 0.5em minus 0.4em\relax IEEE, 2020.

\bibitem{parys2016robust}
B.~P.~G. {Van Parys}, D.~{Kuhn}, P.~J. {Goulart}, and M.~{Morari},
  ``Distributionally robust control of constrained stochastic systems,''
  \emph{IEEE Transactions on Automatic Control}, vol.~61, no.~2, pp. 430--442,
  2016.

\bibitem{samuelson2017data}
S.~{Samuelson} and I.~{Yang}, ``Data-driven distributionally robust control of
  energy storage to manage wind power fluctuations,'' in \emph{2017 IEEE
  Conference on Control Technology and Applications (CCTA)}, 2017, pp.
  199--204.

\bibitem{sinha2020formulazero}
A.~Sinha, M.~O'Kelly, H.~Zheng, R.~Mangharam, J.~Duchi, and R.~Tedrake,
  ``Formulazero: Distributionally robust online adaptation via offline
  population synthesis,'' \emph{arXiv preprint arXiv:2003.03900}, 2020.

\bibitem{hakobyan2020wasserstein}
A.~{Hakobyan} and I.~{Yang}, ``Wasserstein distributionally robust motion
  planning and control with safety constraints using conditional
  value-at-risk,'' in \emph{2020 IEEE International Conference on Robotics and
  Automation (ICRA)}, 2020, pp. 490--496.

\bibitem{jacobson1973optimal}
D.~Jacobson, ``Optimal stochastic linear systems with exponential performance
  criteria and their relation to deterministic differential games,'' \emph{IEEE
  Transactions on Automatic control}, vol.~18, no.~2, pp. 124--131, 1973.

\bibitem{whittle1981risk}
P.~Whittle, ``Risk-sensitive linear/quadratic/gaussian control,''
  \emph{Advances in Applied Probability}, vol.~13, no.~4, pp. 764--777, 1981.

\bibitem{exarchos2016game}
I.~Exarchos, E.~A. Theodorou, and P.~Tsiotras, ``Game-theoretic and
  risk-sensitive stochastic optimal control via forward and backward stochastic
  differential equations,'' in \emph{2016 IEEE 55th Conference on Decision and
  Control (CDC)}.\hskip 1em plus 0.5em minus 0.4em\relax IEEE, 2016, pp.
  6154--6160.

\bibitem{wang2020game}
M.~Wang, N.~Mehr, A.~Gaidon, and M.~Schwager, ``Game-theoretic planning for
  risk-aware interactive agents,'' in \emph{2020 IEEE/RSJ International
  Conference on Intelligent Robots and Systems}.\hskip 1em plus 0.5em minus
  0.4em\relax IEEE, 2020.

\bibitem{bertsekas1976dynamic}
D.~P. Bertsekas, \emph{Dynamic Programming and Stochastic Control}.\hskip 1em
  plus 0.5em minus 0.4em\relax USA: Academic Press, Inc., 1976.

\bibitem{astrom1970introduction}
K.~J. {\AA}str{\"o}m, \emph{Introduction to stochastic control theory}.\hskip
  1em plus 0.5em minus 0.4em\relax Academic Press, 1970.

\bibitem{jacobson1970differential}
D.~H. Jacobson and D.~Q. Mayne, \emph{Differential dynamic programming}.\hskip
  1em plus 0.5em minus 0.4em\relax North-Holland, 1970.

\bibitem{li2004iterative}
W.~Li and E.~Todorov, ``Iterative linear quadratic regulator design for
  nonlinear biological movement systems.'' in \emph{ICINCO (1)}, 2004, pp.
  222--229.

\bibitem{todorov2005generalized}
E.~Todorov and W.~Li, ``A generalized iterative lqg method for locally-optimal
  feedback control of constrained nonlinear stochastic systems,'' in
  \emph{Proceedings of the 2005, American Control Conference, 2005.}\hskip 1em
  plus 0.5em minus 0.4em\relax IEEE, 2005, pp. 300--306.

\bibitem{tassa2012synthesis}
Y.~Tassa, T.~Erez, and E.~Todorov, ``Synthesis and stabilization of complex
  behaviors through online trajectory optimization,'' in \emph{2012 IEEE/RSJ
  International Conference on Intelligent Robots and Systems}.\hskip 1em plus
  0.5em minus 0.4em\relax IEEE, 2012, pp. 4906--4913.

\bibitem{farshidian2015risk}
F.~Farshidian and J.~Buchli, ``Risk sensitive, nonlinear optimal control:
  Iterative linear exponential-quadratic optimal control with gaussian noise,''
  \emph{arXiv preprint arXiv:1512.07173}, 2015.

\bibitem{roulet2020convergence}
V.~Roulet, M.~Fazel, S.~Srinivasa, and Z.~Harchaoui, ``On the convergence of
  the iterative linear exponential quadratic gaussian algorithm to stationary
  points,'' in \emph{2020 American Control Conference (ACC)}.\hskip 1em plus
  0.5em minus 0.4em\relax IEEE, 2020, pp. 132--137.

\bibitem{doyle1978guaranteed}
J.~C. Doyle, ``Guaranteed margins for lqg regulators,'' \emph{IEEE Transactions
  on automatic Control}, vol.~23, no.~4, pp. 756--757, 1978.

\bibitem{dupuis2011weak}
P.~Dupuis and R.~S. Ellis, \emph{A weak convergence approach to the theory of
  large deviations}.\hskip 1em plus 0.5em minus 0.4em\relax John Wiley \& Sons,
  1997.

\bibitem{van2012motion}
J.~Van Den~Berg, S.~Patil, and R.~Alterovitz, ``Motion planning under
  uncertainty using iterative local optimization in belief space,'' \emph{The
  International Journal of Robotics Research}, vol.~31, no.~11, pp. 1263--1278,
  2012.

\bibitem{rubinstein2013cross}
R.~Y. Rubinstein and D.~P. Kroese, \emph{The cross-entropy method: a unified
  approach to combinatorial optimization, Monte-Carlo simulation and machine
  learning}.\hskip 1em plus 0.5em minus 0.4em\relax Springer Science \&
  Business Media, 2013.

\bibitem{kochenderfer2019algorithms}
M.~J. Kochenderfer and T.~A. Wheeler, \emph{Algorithms for optimization}.\hskip
  1em plus 0.5em minus 0.4em\relax Mit Press, 2019.

\bibitem{kusumoto2019informed}
R.~Kusumoto, L.~Palmieri, M.~Spies, A.~Csiszar, and K.~O. Arras, ``Informed
  information theoretic model predictive control,'' in \emph{2019 International
  Conference on Robotics and Automation (ICRA)}.\hskip 1em plus 0.5em minus
  0.4em\relax IEEE, 2019, pp. 2047--2053.

\bibitem{schmerling2018multimodal}
E.~Schmerling, K.~Leung, W.~Vollprecht, and M.~Pavone, ``Multimodal
  probabilistic model-based planning for human-robot interaction,'' in
  \emph{2018 IEEE International Conference on Robotics and Automation
  (ICRA)}.\hskip 1em plus 0.5em minus 0.4em\relax IEEE, 2018, pp. 1--9.

\bibitem{ivanovic2020mats}
B.~Ivanovic, A.~Elhafsi, G.~Rosman, A.~Gaidon, and M.~Pavone, ``Mats: An
  interpretable trajectory forecasting representation for planning and
  control,'' \emph{arXiv preprint arXiv:2009.07517}, 2020.

\bibitem{chai2019multipath}
Y.~Chai, B.~Sapp, M.~Bansal, and D.~Anguelov, ``Multipath: Multiple
  probabilistic anchor trajectory hypotheses for behavior prediction,'' in
  \emph{Conference on Robot Learning}, 2019, pp. 86--99.

\bibitem{wang2020fast}
A.~Wang, X.~Huang, A.~Jasour, and B.~Williams, ``Fast risk assessment for
  autonomous vehicles using learned models of agent futures,'' in
  \emph{Robotics: Science and Systems 2020}, 2020.

\bibitem{salzmannivanovic2020trajectron++}
T.~Salzmann, B.~Ivanovic, P.~Chakravarty, and M.~Pavone, ``Trajectron++:
  Dynamically-feasible trajectory forecasting with heterogeneous data,'' in
  \emph{{European Conf. on Computer Vision}}, Aug. 2020.

\bibitem{kapteyn2018distributionally}
M.~G. Kapteyn, K.~E. Willcox, and A.~Philpott, ``A distributionally robust
  approach to black-box optimization,'' in \emph{2018 AIAA Non-Deterministic
  Approaches Conference}, 2018, p. 0666.

\bibitem{belghazi2018mine}
M.~I. Belghazi, A.~Baratin, S.~Rajeswar, S.~Ozair, Y.~Bengio, A.~Courville, and
  R.~D. Hjelm, ``Mine: mutual information neural estimation,'' \emph{arXiv
  preprint arXiv:1801.04062}, 2018.

\bibitem{iranzad2019estimation}
M.~N. Iranzad, ``Estimation of information measures and its applications in
  machine learning,'' Ph.D. dissertation, University of Michigan, 2019.

\bibitem{chua2018deep}
K.~Chua, R.~Calandra, R.~McAllister, and S.~Levine, ``Deep reinforcement
  learning in a handful of trials using probabilistic dynamics models,'' in
  \emph{Advances in Neural Information Processing Systems}, 2018, pp.
  4754--4765.

\end{thebibliography}
